\pdfoutput=1

\documentclass[11pt]{article}
\usepackage{svg}
\usepackage{booktabs}
\usepackage{multirow}
\usepackage[normalem]{ulem}
\usepackage{amsmath}

\usepackage[]{ACL2023}

\usepackage{times}
\usepackage{latexsym}
\usepackage{float}
\usepackage{color,soul}
\usepackage{todonotes}
\usepackage{subfig}

\usepackage[T1]{fontenc}

\usepackage[utf8]{inputenc}

\usepackage{microtype}

\usepackage{inconsolata}

\title{From Chatter to Matter: Addressing Critical Steps of\\Emotion Recognition Learning in Task-oriented Dialogue}

\author{Shutong Feng, Nurul Lubis, Benjamin Ruppik, Christian Geishauser, Michael Heck, \\ {\bf Hsien-chin Lin, Carel van Niekerk, Renato Vukovic, and Milica Ga\v{s}i\'{c}} \\
  Heinrich Heine University Düsseldorf, Germany\\
  \texttt{\{fengs,lubis,ruppik,geishaus,heckmi,linh,niekerk,revuk100,gasic\}@hhu.de} \\}

\begin{document}
\setlength{\abovedisplayskip}{4pt}
\setlength{\belowdisplayskip}{4pt}

\maketitle
\begin{abstract}

Emotion recognition in conversations (ERC) is a crucial task for building human-like conversational agents. While substantial efforts have been devoted to ERC for chit-chat dialogues, the task-oriented counterpart is largely left unattended. Directly applying chit-chat ERC models to task-oriented dialogues (ToDs) results in suboptimal performance as these models overlook key features such as the correlation between emotions and task completion in ToDs. In this paper, we propose a framework that turns a chit-chat ERC model into a task-oriented one, addressing three critical aspects: data, features and objective. 
First, we devise two ways of augmenting rare emotions to improve ERC performance.
Second, we use dialogue states as auxiliary features to incorporate key information from the goal of the user. Lastly, we leverage a multi-aspect emotion definition in ToDs to devise a multi-task learning objective and a novel emotion-distance weighted loss function. Our framework yields significant improvements for a range of chit-chat ERC models on EmoWOZ, a large-scale dataset for user emotion in ToDs. We further investigate the generalisability of the best resulting model to predict user satisfaction in different ToD datasets. A comparison with supervised baselines shows a strong zero-shot capability, highlighting the potential usage of our framework in wider scenarios.

\end{abstract}

\section{Introduction}
\label{sec:introduction}

Emotion recognition in conversations (ERC) is a crucial task in conversational artificial intelligence research because it lays the foundation for affective abilities in computers such as empathetic response generation \citep{Picard97}. Over years, it has shown values in downstream applications such as opinion mining \citep{Colneric2020EmotionRO} and human-like dialogue modelling \citep{Zhou_Huang_Zhang_Zhu_Liu_2018}.

Dialogue systems can be broadly categorised into two categories: (1) chit-chat or open-domain systems and (2) task-oriented dialogue (ToD) systems. Chit-chat systems are set up to mimic human behaviours in a conversation \citep{10.5555/1214993}. There are no particular goals associated with the dialogue and the system aims to keep the user engaged with natural and coherent responses. On the other hand, ToD systems are concerned with fulfilling user goals, such as information retrieval for hotel booking \citep{young2002talking}.

Recently, the difference between chit-chat and ToD systems have been blurred by the utilisation of pre-trained language models as back-bone to both types of systems. However, emotions in ToDs and chit-chat dialogues play different roles and are therefore expressed differently \citep{feng-etal-2022-emowoz}. This highlights the need for dedicated emotion modelling methods for each system. 

\begin{figure}
\centering
\subfloat[Chit-chat dialogue from \citet{li-etal-2017-dailydialog}]{\includegraphics[width=\columnwidth]{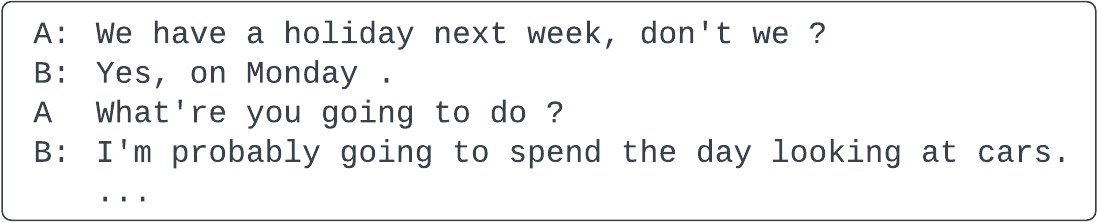}}\qquad
\subfloat[Task-oriented dialogue from \citet{budzianowski-etal-2018-multiwoz}]{\includegraphics[width=\columnwidth]{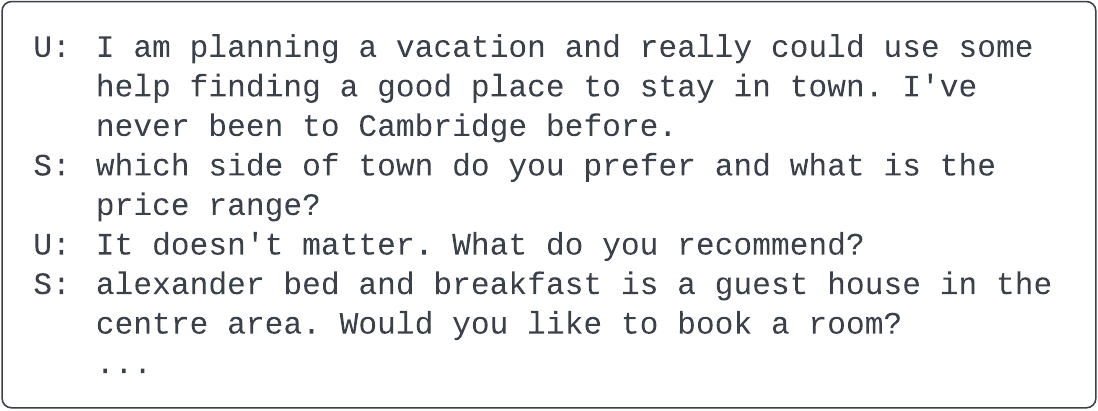}}\qquad
\caption{Comparison of dialogues about holiday in chit-chat dialogues and task-oriented dialogues.\label{fig:dialogue-comparison}}
\end{figure}

As illustrated in Figure \ref{fig:dialogue-comparison}, in chit-chat dialogues, speakers make use of emotions to facilitate communication by, for example, raising empathy as a result of emotion-eliciting situations or topics. On the other hand, emotions in ToDs are centred around the user's goal, and therefore emotion cues lie in both the user's wording and the task performance.

While many large-scale corpora for emotions in chit-chat dialogues exist \citep{Busso2008IEMOCAPIE,5959155,lubis2015construction,li-etal-2017-dailydialog,zahiri:18a}, there are considerably fewer resources for emotions in ToDs. EmoWOZ, which evolved from MultiWOZ, a widely used ToD dataset, is one notable exception \citep{feng-etal-2022-emowoz}. It contains a novel emotion description that is designed for ToDs and inspired by the Ortony-Clore-Collins (OCC) model \citep{ortony_clore_collins_1988}. Emotion is described in terms of three aspects: \textbf{valenced} (positive or negative) reactions towards \textbf{elicitors} (operator, user, or event) in a certain \textbf{conduct} (polite or impolite). However, due to the nature of ToDs, the occurrence of some emotions (e.g. users expressing feelings about their situations) are very rare, leading to a class imbalance in the corpus.

Similarly, advancements on the ERC task are mainly focused on chit-chat dialogues, involving an array of diverse factors from speaker personality \citep{Majumder_Poria_Hazarika_Mihalcea_Gelbukh_Cambria_2019} to commonsense knowledge \citep{ghosal-etal-2020-cosmic}.
Nevertheless, since these models are designed for chit-chat dialogues, they overlook how emotions are triggered and expressed with respect to goal completion in task-oriented context. The work of \citet{1221370} is among one of the earliest and very few to address emotion detection in ToDs but uses generic unigram models instead of dedicated approaches. 

In this work, we tackle critical steps of ERC in ToDs from three angles: the data, the features, and the learning objective. In particular,
\begin{description}
\setlength\itemsep{0em}
    \item[Data:] we address the poor ERC performance of particularly rare emotions in ToDs via two strategies of data augmentation (DA),
    \item[Features:] we leverage dialogue state information and sentiment-aware textual features,
    \item[Objective:] we exploit the three aspects of emotions, namely valence, elicitor, and conduct, in two ways: as a multi-task learning (MTL) objective and to define a novel emotion-distance-weighted loss (\textit{EmoDistLoss}).
\end{description}

To the best of our knowledge, our work is the first to provide dedicated methods for emotion recognition in ToDs. Our experiments and analyses show that our framework leads to significant improvements for a range of chit-chat ERC models when evaluated on EmoWOZ. 

We further investigate the generalisability of the best resulting model to predict user satisfaction in various ToD datasets under zero-shot transfer. Our model achieves comparable results as supervised baselines, demonstrating strong zero-shot capability and potential to be applied in wider scenarios.

\section{Related Work}
\subsection{ERC Datasets}
Early work on ERC relied on small scale datasets \citep{Busso2008IEMOCAPIE,5959155,lubis2015construction}. More recently, a few large-scale datasets have been made available to the research community. They contain dialogues from emotion-rich and spontaneous scenarios such as daily communications \citep{li-etal-2017-dailydialog} and situation comedies \citep{zahiri:18a}. 

For ToDs, the majority of available datasets address only one particular aspect of emotions such as sentiment polarity \citep{10.1371/journal.pone.0235367,shi-yu-2018-sentiment}, user satisfaction \citep{schmitt-etal-2012-parameterized,10.1145/3404835.3463241}, and politeness \citep{https://doi.org/10.48550/arxiv.2210.12942,10.1371/journal.pone.0278323}. For more fine-grained emotions, \citet{singh-etal-2022-emoinhindi} constructed EmoInHindi for emotion category and intensity recognition in mental health and legal counselling dialogues in Hindi, and \citet{feng-etal-2022-emowoz} released EmoWOZ, which concerns user emotions in human-human and human-machine in information-seeking dialogues. Among these datasets, EmoWOZ has the largest scale, accompanied with a label set tailored to the task-oriented scenario.

\subsection{Data Augmentation (DA)} DA is an effective approach to improve model performance
by improving data diversity without explicitly collecting more data. 
While textual DA can be performed in the feature space via interpolation and sampling \citep{kumar2019closer}, it is commonly performed in the data space for controllability. Rule-based methods involve operations such as insertion and substitution \citep{wei-zou-2019-eda}. While they are easy to implement, the diversity in augmented samples depends on the complexity of the rules. On the contrary, model-based methods are more scalable. 
These typically include the use of language models \citep{jiao-etal-2020-tinybert}, 
translation models \citep{10.5555/3495724.3496249}, and paraphrasing methods \citep{hou-etal-2018-sequence}.


Additional training samples can also be obtained from unlabelled data via weak supervision \citep{Ratner_2017}. To generate the automatic labels, a single model or an ensemble of models may be used. This method can be interpreted as self-augmentation \citep{Xu_Zhang_He_Ge_Yang_Yang_Wu_2022}, self-training \citep{Xie_2020_CVPR}, or distillation \citep{Radosavovic2017DataDT}.

DA has also been also deployed in ToD modelling. \citet{hou-etal-2018-sequence} generated samples by paraphrasing delexicalised utterances. \citet{gritta-etal-2021-conversation} conceptualised ToDs into transitional graphs and generate new dialogue paths by sampling. \citet{10.1162/tacl_a_00513} proposed a weak supervision framework to address the lack of fine-grained span labels for dialogue state tracking. DA for emotions in ToDs requires careful considerations to avoid emotion mismatch and is not yet explored.

\subsection{ERC Models and Features}
Text-based ERC is in essence a text classification problem with an emphasis on contextual modelling. \citet{poria-etal-2017-context} proposed a recurrent neural network (RNN) for multimodal ERC. The follow-up work of \citet{Majumder_Poria_Hazarika_Mihalcea_Gelbukh_Cambria_2019} considered speaker-specific context. 
ERC performance has been continuously improved by techniques such as incorporating external knowledge \citep{ghosal-etal-2020-cosmic} and contrastive learning \citep{song-etal-2022-supervised}.

\paragraph{Sentiment-aware Embeddings} 
Word-vector embeddings tailored for a particular natural language processing task can effectively improve the performance for that task~\cite{10.1145/3434237}. In a similar vein, \citet{tang-etal-2014-learning} incorporated sentiment classification objectives in the training of the word embedding model of \citet{Collobert2008AUA} specifically for sentiment analysis. 
\citet{yu-etal-2017-refining} refined static word embeddings with the aid of a sentiment lexicon. Later, many sentiment-aware variants of pre-trained language models were obtained by incorporating sentiment-related objectives in training \citep{xu-etal-2019-bert,yin-etal-2020-sentibert,zhou-etal-2020-sentix}. They successively achieved state-of-the-art performance in sentiment analysis tasks among language representation models.
 
\subsection{Learning Objectives for ERC Models}
ERC is often considered a single-label sequential classification problem. Using softmax cross-entropy loss has been the norm in the training of deep learning ERC models for categorical emotions \citep{poria-etal-2017-context,zhong-etal-2019-knowledge,ghosal-etal-2020-cosmic,kim2021emoberta} or quantised emotion dimensions \citep{cerisara-etal-2018-multi,Wang_Wang_Sun_Li_Liu_Si_Zhang_Zhou_2020}. However, this simplistic cross-entropy loss ignores the inter-class relations and output probabilities on incorrect classes. 

\citet{chen2018complement} proposed to suppress the output probabilities of incorrect classes equally while minimising the standard cross-entropy loss. \citet{Hou2016SquaredEM} proposed squared earth mover’s distance to penalise the misclassifications according to a ground distance matrix that quantifies the dissimilarities between classes for image age estimation and aesthetics estimation. 

Although highly suitable for emotions, learning from misclassifications is rarely considered because the distance between emotion classes is hard to quantify. Therefore, we propose to leverage the structured label definition of EmoWOZ to model inter-class similarity. 

\paragraph{Multi-task Learning (MTL)} is a technique for learning tasks in parallel using a shared representation. It aims to improve generalisation by using the information in training signals of related tasks as an inductive bias \citep{Caruana1997}. In emotion recognition, auxiliary tasks include topic classification \citep{Wang_Wang_Sun_Li_Liu_Si_Zhang_Zhou_2020} and personality traits \citep{https://doi.org/10.48550/arxiv.2101.02346}. When co-labels are not available, it is also possible to leverage aspects of emotion for additional labels such as valence-arousal \citep{8282123}. In this work, we exploit the valence-elicitor-conduct labels in EmoWOZ for MTL.

\section{Background}

\subsection{User Emotion Recognition} 
\label{sec:problem-definition}
We formulate the task as recognising one emotion class $ e_t $ from a set of $n$ discrete emotions $E=\{e^1, e^2, ..., e^n\}$ in the user turn $u_t$, given a dialogue history $H_t = [u_t, s_{t-1}, u_{t-1}, ..., s_1, u_1]$, where $s$ denotes system turns and $u$ denotes user turns. 
Unlike existing chit-chat ERC models, which are often built for static analysis on the dialogue as a whole, real-time ERC in ToDs does not consider future utterances in dialogue.

\subsection{User Satisfaction Prediction}

User satisfaction prediction aims to predict one satisfaction level $c_t$ from a set of $m$ discrete levels $C=\{c^1, c^2, ..., c^m\}$ in the user turn $u_t$, given all previous turns $P_t = [s_{t-1}, H_{t-1}]$. This task differs from ERC in that the user turn $u_t$ is not available as a part of model input. Since user satisfaction is highly correlated with the valence aspect in user emotion, this task can also be viewed as user emotion prediction. This is an important task in building ToD systems and has been used for user simulation and system evaluation \citep{10.1145/3404835.3463241}. 



\section{\textbf{E}motion \textbf{R}ecogniser for \textbf{T}ask-\textbf{o}riented \textbf{D}ialogues (ERToD)}
\begin{figure}[h]
\centering
\includegraphics[width=\columnwidth]{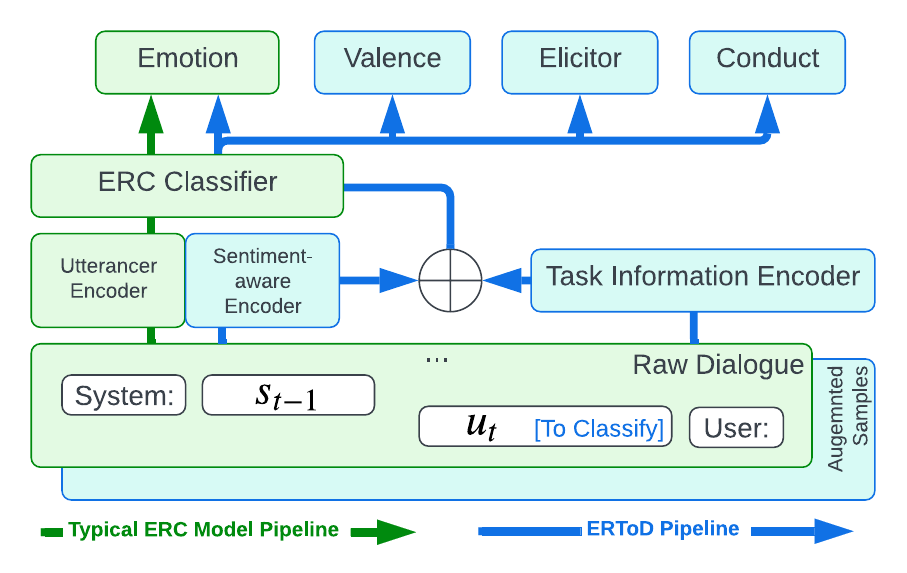}
\caption{Our proposed ERToD Framework.\label{fig:model}}
\end{figure}

In this section, we propose our ERToD framework that adapt chit-chat ERC models to the task-oriented domain, as illustrated in Figure \ref{fig:model}.


\subsection{Data Augmentation}
Unlike emotions in chit-chat dialogues, resources for emotions in ToDs are very limited. In addition, the data scarcity not only lies in the lack of linguistic diversity but also in the limited domains and actions in which emotions are expressed. 

In ToDs, user's emotional expressions have different degrees of connection to the dialogue task. For example, a user can express dissatisfaction towards the system by pointing out the system's mistake. In such a case, simply replacing or paraphrasing the user's utterance based on emotion can potentially break the consistency of the task flow in the context. Such emotions are \textit{context-dependent}.


On the other hand, \textit{context-independent emotions} are expressed without any connection to the user goal, such is the case with abusive utterances. Due to the lack of connection, a simple replacement with a different abusive sentence can fit into the context well without impairing the consistency of task flow in the dialogue.

To obtain augmented samples with meaningful and coherent context, we adopt two different strategies of DA according to the degree of context dependency of emotional expressions.

\paragraph{Context-independent Emotions} 
To augment samples for a target emotion $e$, we select a user utterance $u'$ with the equivalent label from other dialogue datasets. We then use it to replace the user utterance $u_t$ having label $e$ in the training data while keeping the original context $[s_{t-1}, u_{t-1}, ..., s_1, u_1]$. The new sample is obtained as $H'_t = [u', s_{t-1}, u_{t-1}, ..., s_1, u_1]$.

\paragraph{Context-dependent Emotions} We first sample a pool of unlabelled candidate dialogues $H'_t = [u'_t, s'_{t-1}, u'_{t-1}, ..., s'_1, u'_1]$ from other ToD datasets. 
We train a classifier with an uncertainty estimator to identify the emotion label $e_t$ of the user utterance $u_t$ and its confidence in each candidate: 
\begin{equation}
p(e_t), \textrm{conf}(e_t) = \textrm{UncertaintyClassifier}(H'_t)    
\end{equation}
The candidate is selected for emotion $e_t$ only if $\textrm{conf}_t(e)$ is above a confidence threshold $\theta$. 

\subsection{Task Information Encoder}

We use a dialogue state tracker (DST) to determine the status of goal completion at each turn. In ToDs, the dialogue state describes the system's understanding of the  the user's goal up to that point in the dialogue \citep{YOUNG2010150}. It encodes dialogue progress in an abstractive manner.

Here as a proof of concept, we use an ontology-dependent DST, which means the concepts that the system can talk about are pre-determined. While we can eliminate the ontology dependency by, for example, using an ontology-independent DST and extracting task features from dialogue state description in natural language, this goes beyond the scope of this work. The DST takes the dialogue history to determine $\textrm{SemDS}_t$, the current dialogue state in semantic form. It is stored as a dictionary that records slots and filled values.
$\textrm{SemDS}_t$ is then converted into a vector of 0/1's, indicating whether a particular slot has been filled. 
\begin{equation}
{V}_{t} = \textrm{Vectoriser(SemDS}_t\textrm{)}    
\end{equation}
To account for the change of dialogue state, which depicts how the system performs locally, we concatenate dialogue states of three consecutive turns to obtain a contextual dialogue state vector.
\begin{equation}
\label{eqn:dialog-state-vector}
    \widetilde{V}_{t} = {V}_t \oplus {V}_{t-1} \oplus {V}_{t-2}
\end{equation}
$\textrm{V}_{t \leq 0}$ are zero vectors, representing the state before the dialogue starts. $\widetilde{V}_{t}$ is then fed into a trainable fully connected (FC) layer.
\begin{equation}
    {S}_t = \textrm{FC}(\widetilde{V}_{t})
    \label{eqn:fc-layer}
\end{equation}

\paragraph{Feature Fusion for Emotion Classification} For a chit-chat ERC model with an arbitrary utterance encoder, $R_t = \textrm{Encoder(}H_t\textrm{)}$, i.e. $R_t$ is the encoded representation of the dialogue history $H_t$. 
The utterance encoder is replaced with a sentiment-aware encoder in our framework~(see Figure~\ref{fig:model}).

The utterance and the task information encodings are fused via concatenation and fed into the emotion classifier. The output probability of all emotion classes in utterance $u_t$ is given by:
\begin{equation}
    p_t = \textrm{Softmax}(\textrm{Classifier}(R_{t} \oplus {S}_t))
\end{equation}

\subsection{Learning Objectives}
\subsubsection{Emotion-Distance Weighted Loss}
Emotion classification is a very challenging task due to the subjectivity in the perception of emotion. Since some emotions are more similar to each other than others, it may be advantageous to distinguish marginally wrong recognitions (satisfied vs excited) from extremely wrong ones (satisfied vs dissatisfied). Furthermore, different misclassifications can elicit different user reactions to the dialogue agent. 
For example, perceiving satisfaction when the user is neutral may or may not annoy the user, but accusing the user of abusive behavior by mistake is a serious offense to the user.
Therefore, it is intuitive to penalise misclassifications according to (1) the distance 
from the label and (2) output probabilities on incorrect labels.

\paragraph{Defining the Emotion Distance}
Since emotion labels in EmoWOZ are defined in three aspects, we can define the distance between emotion labels in terms of their distance on each aspect. A matrix $D$ is defined where each element $D(i,j)$ is a vector containing the distance between emotion label $i$ and $j$ in each of three aspects (valence, elicitor, and conduct). 
The matrix $D$ is symmetric with vector-valued entries.
\begin{equation}
\label{eqn:dist-definition}
    D(i,j) = [d_{val}(i,j), d_{eli}(i,j), d_{con}(i,j)]
\end{equation}
The final distance is obtained by the sum of the distance in each aspect, followed by an addition of 1 and smoothing with the log operator. The addition of 1 ensures that the log distance is still 0 for identical labels.
\begin{equation}
\label{eqn:norm-dist}
    \widetilde{D}(i,j) = \log{(\textrm{sum}(D(i,j)) + 1)}
\end{equation}

\paragraph{Considering Misclassification Probabilities}
For each sample including the dialogue history $H_t$, we look at the softmax output from the model.
\begin{equation}
    p_t = \textrm{Classifier}(H_t)
\end{equation}

We aim to minimise the probability of each misclassification $p_t(e=e_i)$ where $e_i \neq \textrm{label}_t$. This is done by maximising $1 - p_t(e=e_i)$, the probability of the utterance \textit{not} being wrongly recognised as $e_i$. 
We then calculate the log of this probability so that in the case of a perfectly correct recognition, the penalty from misclassification will be 0.
\begin{equation}
\label{eqn:log-terms}
    f(p_{t}) = \log{(1 - p_t)}
\end{equation}

\paragraph{Obtaining Weights for Misclassifications}
We obtain the relevant row in matrix $D$ that contains the distance between each emotion and the ground-truth label $j$ of utterance $u_t$, followed by a normalisation to obtain a vector $w_{t,j}$ of normalised emotion-distance weights for all emotions.
\begin{equation}
    o_{t,j} = \textrm{onehot}(\textrm{label}_t=j)
\end{equation}
\begin{equation}
    \widetilde{D}(:,j) = \widetilde{D} \times o_{t,j}
\end{equation}
\begin{equation}
    w_{t,j} = \widetilde{D}(:,j) / \textrm{sum}(\widetilde{D}(:,j))
\end{equation}


\paragraph{EmoDistLoss} The final loss, which we name \emph{EmoDistLoss}, is calculated from the negative weighted sum of log terms from Equation \ref{eqn:log-terms}. Since the distance, hence the weight, between identical labels is 0, this calculation does not involve the output probability of the correct label.
\begin{equation}
    \textrm{\textit{EmoDistLoss}}_{t} = - w_{t,j} \cdot f(p_t)
\end{equation}

\subsubsection{MTL via Emotional Aspects}
In addition to the emotion classification head, we have a classification head for each emotion aspect from the label definition, namely the valence, the elicitor, and the conduct.

The overall classification loss $L$ is a weighted sum of the loss from softmax outputs of four classification heads $L_{emo}, L_{val}, L_{eli}, L_{con}$ with a hyperparameter $\alpha$.
\begin{equation}
    L = \alpha L_{emo} + \frac{1}{3}(1-\alpha)(L_{val} + L_{eli} + L_{con})
    \label{eqn:weighted-loss}
\end{equation}

\section{User Emotion Recognition in ToDs with ERToD}
\subsection{Experimental Set-up}
\subsubsection{Dataset}
\label{sec:dataset}
We train and test our models on EmoWOZ. It contains user emotion annotations for all dialogues from MultiWOZ \citep{budzianowski-etal-2018-multiwoz} and additional 1000 human-machine dialogues. 
It contains 7 emotion groups (see Table \ref{tab:emowoz} and Appendix \ref{sec:emotion-definition} for details). Four emotion classes are considerably rare: \textit{fearful}, \textit{apologetic}, \textit{abusive}, and \textit{excited}. DA examples can be found in Appendix \ref{sec:da-examples}. Our primary aim of DA is to address the poor ERC performance on rare emotions rather than building a balanced dataset. While the later aim can be achieved with the aid of large language models for example, this is out of the scope of our work.

\begin{table}[htpb]
\centering
\scriptsize
\begin{tabular}{l|ccc|r}
\toprule[1pt]
 \multicolumn{1}{c|}{Class Name} & Valence & Elicitor & Conduct & \multicolumn{1}{c}{Count (\%)} \\ \hline
\textbf{Neutral} & Neutral & Don't Care & Polite & 58,656 (70.1\%) \\
\textbf{Satisfied} & Positive & Operator & Polite & 17,532 (21.0\%) \\
\textbf{Dissatisfied} & Negative & Operator & Polite & 5,117  (6.1\%) \\
\textbf{Excited} & Positive & Event/Fact & Polite & 971  (1.2\%) \\
\textbf{Apologetic} & Negative & User & Polite & 840  (1.0\%) \\
\textbf{Fearful} & Negative & Event/Fact & Polite & 396  (0.5\%) \\
\textbf{Abusive} & Negative & Operator & Impolite & 105  (0.2\%) \\
\bottomrule[1pt]
\end{tabular}
\caption{EmoWOZ Emotion definition and distribution. 
\label{tab:emowoz}}
\vspace{-7mm}
\end{table}

\paragraph{Augmenting Abusive Utterances} The user sometimes becomes abusive towards the system. While this correlates with failure to satisfy the user goal, exact abusive expressions uttered by the user are usually independent of the context. Therefore, we apply our DA method for context-independent emotions for \textit{Abusive}. We utilise ConvAbuse, a dataset for nuanced abusive behaviours in chit-chat conversations \citep{cercas-curry-etal-2021-convabuse}, for more diverse abusive expressions. In ConvAbuse, user utterances are labelled with type, target, strength, and directiveness. We filter for abuses on the system's intellectuality (labelled as \verb|type=intellectual| and \verb|target=system|) to better suit ToD context. We combine each selected utterance with the context of a random abusive utterance in EmoWOZ, resulting in 273 augmented samples. 

\paragraph{Augmenting Fearful, Apologetic, and Excited Utterances} Expressions of these emotions usually contain task information. \textit{Fearful} and \textit{Excited} usually co-occur with a description of the situation that prompts the user to interact with the system. \textit{Apologetic} is frequently associated with a correction of search criteria. There is a strong connection between these emotion expressions and the progression of the task in the dialogue history. Therefore, we apply our DA method for these context-dependent emotions. We look for samples with desired emotions from other ToD datasets using automatic labels. We train a ContextBERT on EmoWOZ (see Section \ref{sec:baselines}) with a 30\% dropout on the BERT output. We train the model with 10 different seeds and run inferences on the training set of existing ToD datasets: Schema-Guided Dialogue (SGD, \citealt{{Rastogi2019TowardsSM}}), Taskmaster-1 (TM-1), and Taskmaster-2 (TM-2) \citep{byrne-etal-2019-taskmaster}. In addition, we filter for common domains of EmoWOZ: \textit{Hotels, RideSharing, Travel, Restaurants} in SGD, \textit{RestaurantTable, PizzaOrdering, CoffeeOrdering, UberLyft} in TM-1, and \textit{HotelSearch, Restaurants, FoodOrdering} in TM-2. The classification confidence is measured by votes from 10 models. We use a confidence threshold of $0.7$ and cap the number of augmented samples at $1000$ for each emotion, resulting in $268$ \textit{fearful}, $872$ \textit{apologetic}, and $1000$ \textit{excited} samples. 




\subsubsection{Baselines}
\label{sec:baselines}
We implement ERToD to a range of ERC models that have been used to benchmark EmoWOZ, as listed in Table \ref{tab:f1-scores}. ContextBERT \citep{feng-etal-2022-emowoz} and EmoBERTa \citep{kim2021emoberta} are simple yet robust transformer-based ERC models, and they have similar spirits except that they respectively use BERT \citep{devlin-etal-2019-bert} and RoBERTa \citep{liu2019roberta} as utterance encoder. They are both built on top of BERT by additionally considering dialogue context and speaker roles in the input. DialogueRNN \citep{Majumder_Poria_Hazarika_Mihalcea_Gelbukh_Cambria_2019} and COSMIC \citep{ghosal-etal-2020-cosmic} are RNN-based models. Following \citep{feng-etal-2022-emowoz}, we use DialogueRNN with either \{GloVe\citep{pennington-etal-2014-glove}+Convolutional Neural Network\} or BERT as the utterance encoder. COSMIC additionally extracts features with a pre-trained commonsense model \citep{bosselut-etal-2019-comet}\footnote{COSMIC requires future utterances in recognising the current emotion whereas other models can be configured as either bidirectional or unidirectional. While we use unidirectional set-ups where possible to comply with our task formulation in Section \ref{sec:problem-definition}, we are also interested in how ERToD improves COSMIC for static dialogue analysis in ToDs.}. 
It is important to note that after replacing the original utterance encoder with the sentiment-aware encoder (as described in Section \ref{sec:training}), two variants of DialogueRNN essentially become the same model, and so do EmoBERTa and ContextBERT.







\subsubsection{Training}
\label{sec:training}
In our task information encoder, we use SetSUMBT DST \citep{van-niekerk-etal-2021-uncertainty} from ConvLab-3 toolkit \citep{https://doi.org/10.48550/arxiv.2211.17148}. SetSUMBT is a strong DST considering uncertainty with a joint goal accuracy of $52.26\%$ on MultiWOZ 2.1 \citep{eric-etal-2020-multiwoz}. The FC layer in Equation \ref{eqn:fc-layer} has input/output dimensions of $1083$ and $256$ respectively and hyperbolic tangent activation (TanH,  \citealt{lecun2015deep}). We further replace the utterance encoders of chit-chat ERC models with SentiX, a sentiment-adapted BERT \citep{zhou-etal-2020-sentix}. 

We use our proposed \textit{EmoDistLoss} for the emotion classification head and cross-entropy loss for MTL heads (valence, elicitor, and conduct). Since the elicitor of \textit{Neutral} emotion is not distinguishable and therefore not explicitly defined in EmoWOZ, we mark the elicitor of \textit{Neutral} samples as \textit{don't care}, and their loss in from elicitor classification is ignored. $\alpha$ in Equation \ref{eqn:weighted-loss} is set to $0.4$ based on several rounds of hyperparameter tuning.

To calculate the \textit{EmoDistLoss}, we use $1$ as the unit distance and define the distance for each emotional aspect as illustrated in Appendix \ref{sec:emotion-distance-definition}. For valence, it is commonly adopted to consider negative and positive as two polarities and neutral in the middle \citep{socher-etal-2013-recursive}. Therefore, the distance is $2$ between positive and negative, and $1$ between non-neutral and neutral. For emotion elicitors, we set the distance between \textit{don't care} to any specific elicitor as $0.5$ to penalise a ``lazy'' classifier that wrongly recognises the emotion as neutral. Doing so also results in a consistent shortest distance of $1$ between any pair of specific elicitors.

We follow the default training set-up of each model except for ContextBERT. We reduce the context size of ContextBERT from $512$ to $128$, resulting in stronger performance and faster training.

\subsubsection{Evaluation}
We report F1 for each emotion. For overall performance, we report both macro F1 and weighted F1. Macro F1 considers each emotion equally and reflects the model's ability to recognise rare emotions. Weighted F1 is the weighted sum of F1 scores of each label. Weights are determined by the proportion of each emotion in the dataset. We exclude \textit{Neutral} from calculating the averages as it makes up more than 70\% of labels.

In addition, we also calculate the average emotion distance (AED) between the recognised emotion and the label to quantify how wrong the model is when it misclassifies. The AED of an emotion $e$ is calculated from the average of 
$\widetilde{D}(\textrm{label=}e,\textrm{recognised\_emotion})$
of samples whose label is $e$ (see Equation \ref{eqn:norm-dist}). Lower AED means less severe consequences from mistakes, and is therefore more desirable. 
All experiments are repeated with 10 different seeds.

\subsection{ERC Results}

\begin{table}[htbp]
    \centering
    \scriptsize
    \setlength\tabcolsep{5pt}
    \begin{tabular}{l|rr|rr|rr}
    \toprule
     & \multicolumn{2}{c|}{Base Model} & \multicolumn{2}{c|}{+ ERToD} & \multicolumn{2}{c}{Difference} \\
     & \multicolumn{1}{c}{MF1} & \multicolumn{1}{c|}{WF1} & \multicolumn{1}{c}{MF1} & \multicolumn{1}{c|}{WF1} & \multicolumn{1}{c}{MF1} & \multicolumn{1}{c}{WF1} \\ \midrule
    BERT & 50.1 & 73.5 & 61.4 & 77.3 & +11.3 & +3.8 \\
    DialogueRNN+GloVe & 40.1 & 74.6 & 56.5 & 78.5 & +16.4 & +3.9 \\
    DialogueRNN+BERT & 52.1 & 75.5 & 56.5 & 78.5 & +4.4 & +3.0 \\
    COSMIC & 56.3 & 77.1 & 57.4 & 79.6 & +1.1 & +2.5 \\
    EmoBERTa & 57.9 & \textbf{83.0} & \textbf{65.9} & \textbf{83.9} & +9.0 & +0.9 \\
    ContextBERT & \textbf{59.1} & 81.9 & \textbf{65.9} & \textbf{83.9} & +6.8 & +2.0 \\
    \bottomrule
    \end{tabular}
 \caption{Macro- and weighted-average F1 (MF1, WF1) of ERC models before and after incorporating ERToD. Best average F1s are marked in \textbf{bold}. All differences are significant with $p<0.05$.
\label{tab:f1-scores}}
\vspace{-3mm}
\end{table}

Table \ref{tab:f1-scores} shows the change in the emotion recognition performance of the selected chit-chat ERC models after incorporating our ERToD framework. ERToD achieves significant improvement in average F1 scores of all models (see Appendix \ref{sec:model-predictions} for examples of model outputs, Appendix \ref{sec:f1-all-emotions} for F1 of individual emotions).

\subsection{Ablation Study on ERToD}

\begin{table}[h]
\centering
\scriptsize
\setlength\tabcolsep{2pt}
\begin{tabular}{c|l|r|r|r|r|r|r|r}
\toprule[1pt]
 & \multicolumn{1}{c|}{Model} & \multicolumn{1}{c|}{Neu.} & \multicolumn{1}{c|}{Sat.} & \multicolumn{1}{c|}{Dis.} & \multicolumn{1}{c|}{Exc.} & \multicolumn{1}{c|}{Apo.} & \multicolumn{1}{c|}{Fea.} & \multicolumn{1}{c}{Abu.}  \\ \midrule
\multicolumn{1}{l|}{\multirow{6}{*}{\rotatebox[origin=c]{90}{F1 Score ($\uparrow$)}}} & ContextBERT & 93.5 & 89.1 & 69.7 & 45.6 & 69.6 & 33.3 & 47.0 \\ 
\multicolumn{1}{l|}{} &  \ + DA & $\dagger$\textbf{94.2} & $\dagger$90.5 & $\dagger$71.0 & 45.3 & $\dagger$72.1 & $\ddagger$38.3 & $\dagger$67.4 \\ 
\multicolumn{1}{l|}{} &  \ \ + DS & $\dagger$\textbf{94.2} & $\dagger$90.5 & $\dagger$71.3 & 45.7 & $\dagger$72.7 & 35.3 & $\dagger$69.4 \\
\multicolumn{1}{l|}{} &  \ \ \ + SentiX & $\dagger$\textbf{94.2} & $\dagger$\textbf{90.6} & $\dagger$72.2 & $\ddagger$47.1 & $\dagger$73.2 & $\dagger$39.0 & $\dagger$66.1 \\ 
\multicolumn{1}{l|}{} &  \ \ \ \  + MTL & $\dagger$\textbf{94.2} & $\dagger$90.4 & $\dagger$\textbf{72.3} & $\ddagger$47.2 & $\dagger$\textbf{73.4} & $\dagger$41.0 & $\dagger$67.9 \\
\multicolumn{1}{l|}{} & \ + ERToD & $\dagger$94.1 & $\dagger$\textbf{90.6} & $\dagger$\textbf{72.3} & $\dagger$\textbf{47.6} & $\dagger$72.0 & $\dagger$\textbf{42.4} & $\dagger$\textbf{69.8} \\ \midrule
\multicolumn{1}{l|}{\multirow{6}{*}{\rotatebox[origin=c]{90}{AED Score ($\downarrow$)}}} & ContextBERT & 0.058 & 0.094 & 0.304 & 0.497 & 0.269 & 0.605 & 0.554 \\ 
\multicolumn{1}{l|}{} & \ + DA & $\dagger$\textbf{0.049} & $\dagger$0.080 & 0.312 & 0.493 & $\ddagger$0.292 & 0.593 & $\dagger$0.339  \\ 
\multicolumn{1}{l|}{} & \ \ + DS & $\dagger$0.053 & $\dagger$0.075 & 0.296 & 0.481 & 0.277 & 0.582 & $\dagger$0.300 \\
\multicolumn{1}{l|}{} & \ \ \ + SentiX & $\dagger$0.052 & $\dagger$0.077 & $\ddagger$0.286 & $\dagger$0.454 & 0.287 & 0.596 & $\dagger$0.283  \\ 
\multicolumn{1}{l|}{} & \ \ \ \ + MTL & $\dagger$0.054 & $\dagger$0.075 & $\ddagger$\textbf{0.284} & $\ddagger$0.456 & 0.277 & 0.585 & $\dagger$\textbf{0.258} \\
\multicolumn{1}{l|}{} & \ + ERToD & 0.056 & $\dagger$\textbf{0.070} & 0.296 & $\dagger$\textbf{0.435} & \textbf{0.244} & \textbf{0.571} & $\dagger$0.277 \\
\bottomrule
\end{tabular}

\caption{F1 ($\uparrow$) and AED ($\downarrow$) scores of \textbf{Neu}tral, \textbf{Sat}isfied, \textbf{Dis}satisfied, \textbf{Exc}ited, \textbf{Apo}logetic, \textbf{Fea}rful, and \textbf{Abu}sive. $\dagger$ indicates statistically significant difference with $p<0.05$ and $\ddagger$ indicates $p<0.1$ when comparing with ContextBERT. Best scores are marked in \textbf{bold}. 
\label{tab:ablation}}
\vspace{-4mm}
\end{table}

We perform an ablation study on the best resulting model, ContextBERT-ERToD (Table \ref{tab:ablation}). We add each technique in the order of data-related, feature-related, and loss-related approaches. Averaged scores can be found in Appendix \ref{sec:f1-ablation-average}.

\paragraph{Impact of DA} DA helps improve almost all F1 scores
even with a relatively small number of additional samples. There is a small and insignificant drop in the F1 of \textit{Excited}, which is also frequently confused among human annotators. 
Further work to resolve the ambiguities would be beneficial.

\paragraph{Impact of Dialogue State (+DS)} Adding dialogue state features further improves most other non-neutral emotions. Although it does not bring advantages for the F1 of \textit{Fearful}, the AED of it continues to improve, showing that the system is making less severe mistakes. 

\paragraph{Impact of SentiX} Initialising BERT with SentiX parameters further improves the recognition of all other non-neutral emotions except for \textit{Abusive}. This suggests that the sentiment information encoded in SentiX is useful for resolving ambiguity. We suspect that, while SentiX is good at distinguishing the valence of emotion, its effect is limited for user conduct, the hallmark of \textit{Abusive}.

\paragraph{Impact of MTL} MTL improves F1 for all non-neutral emotions except for \textit{Satisfied}. It also achieves the best AED for \textit{Abusive}. This suggests that MTL heads, especially the conduct classification head, help identify emotions in the simpler valence-elicitor-conduct space. There is a slight drop in the F1 score of \textit{Satisfied}, but it is compensated by the improvement in its AED.

\paragraph{Impact of \textit{EmoDistLoss} (+ERToD)} The final version of the model achieves the best F1 score in \{\textit{Satisfied, Dissatisfied, Excited, Fearful, Abusive}\} and the best AED score in \{\textit{Satisfied, Excited, Apologetic, Fearful}\}, leading to best averaged scores (Table \ref{tab:ablation-average}). This shows penalising misclassifications according to emotion distance, which is only possible thanks to the emotion model, further helps recognise ambiguous emotions.

For the degradation of both scores in \textit{Neutral}, we hypothesise that the model recognises non-neutral emotions more boldly than annotators, who are more cautious about subtle emotional cues.

\section{Zero-shot User Satisfaction Prediction}

\subsection{Experimental Set-up}

\subsubsection{Dataset} We evaluate our model with \textbf{User Satisfaction Simulation} (USS) dataset where user utterances are annotated with 5-level satisfaction ratings \citep{10.1145/3404835.3463241}. Dialogues in USS come from 5 different ToD datasets:

\paragraph{Jing Dong Dialogue Corpus} (JDDC, \citealp{chen-etal-2020-jddc}) is a multi-turn Chinese dialogue dataset for E-commerce customer service. USS contains 54.5k user satisfaction annotations for 3300 dialouges sampled from JDDC. Since JDDC is in Chinese, we translated it into English with Google Translate API first.

\paragraph{Schema-guided Dialogues} (SGD, \citealp{rastogi2020towards}) is a multi-domain, task-oriented conversations between a human and a virtual assistant. These conversations involve interactions with services and APIs spanning 20 domains, such as banks, events, media, calendar, travel, and weather. USS contains 13.8k user satisfaction annotations for 1000 dialogues sampled from SGD. Although we use SGD for DA, our DA samples do not overlap with SGD dialogues in USS.

\paragraph{Recommendation Dialogue} (ReDial, \citealp{li2018conversational}) is an annotated dataset of dialogues, where users recommend movies to each other. USS contains 11.8k user satisfaction annotations for 1000 dialogues sampled from ReDial. 

\paragraph{Coached Conversational Preference Elicitation} (CCPE, \citealp{radlinski-etal-2019-coached}) is a dialogue dataset where the ``
assistant'' is tasked with eliciting the ``user'' preferences about movies collected in the Wizard-of-Oz framework. USS contains 6.8k user satisfaction annotations for 500 dialogues sampled from CCPE.

\paragraph{MultiWOZ} \citep{budzianowski-etal-2018-multiwoz} is a multi-domain task-oriented dialogue dataset collected in the Wizard-of-Oz framework spanning 7 domains such as restaurant, hotel, and attraction. USS contains 12.5k user satisfaction annotations for 1000 dialogues sampled from MultiWOZ. Since we trained our ERC model on EmoWOZ, which was based on MultiWOZ, we excluded it in our evaluation.

\subsubsection{Baselines}
We compare our zero-shot results with supervised models of \citet{10.1145/3404835.3463241} and \citet{10.1145/3477495.3531814}. HiGRU \citep{yang-etal-2016-hierarchical} and BERT \citep{devlin-etal-2019-bert} were the best two models trained by \citet{10.1145/3404835.3463241} to benchmark USS dataset when it was first released. SatAct and SatActUtt are T5-based models \citep{2020t5}. SatAct is trained to predict user satisfaction and user action in a MTL set-up, whereas SatActUtt additionally incorporates user utterance generation. For satisfaction prediction, these models were set up to predict a 5-level rating during training.

These baseline models were trained on each one of the five ToD subsets in USS with a 10-fold cross-validation. Although non-3 ratings were up-sampled by 10 times in their training, the training data size is still smaller than that of ContextBERT-ERToD (68.9k emotion annotations, EmoWOZ and DA samples altogether).

\subsubsection{Zero-shot Inference}
We experimented with ContextBERT-ERToD, the best resulting model from ERC training. After training the model for ERC, we fixed its parameters and ran inference with USS dataset for zero-shot user satisfaction prediction. To adapt to user satisfaction prediction set-up, we excluded information about the user turn at $t$ from the model input as well as the dialogue state. Specifically, for utterance encoding, we excluded $u_t$ from the dialogue history to have $H_t = [s_{t-1}, u_{t-1}, ..., s_1, u_1]$. For task information encoding, we shifted the context window in Equation \ref{eqn:dialog-state-vector} by one and have $\widetilde{V}_{t} = {V}_{t-1} \oplus {V}_{t-2} \oplus {V}_{t-3}$ as the new contextual dialogue state vector. 

\subsubsection{Evaluation}
In the works of baseline models, satisfaction ratings \{1,2\} were considered the negative class and \{3,4,5\} as the positive. To map the emotion prediction from our ERC model to binary satisfaction ratings, it is intuitive to leverage the valence aspect of emotions. Emotion classes with a negative valence were considered \textit{Not Satisfied} and those with a positive valence as \textit{Satisfied}. The emotion \textit{Apologetic} is an exception among emotions with a negative valence. Since its elicitor is the user him/herself, it should not be considered as a sign of user dissatisfaction. Regarding the emotion class \textit{Neutral}, we mapped it to \textit{Satisfied} because the original evaluation set-up of baseline models considered the medium satisfaction rating, 3, as the positive class. 

Overall, we considered \{\textit{Neutral, Apologetic, Excited, Satisfied}\} as the positive class and \{\textit{Fearful, Dissatisfied, Abusive}\} as negative.

\subsection{Results}
\begin{table}[htbp]
    \centering
    \scriptsize
    \setlength\tabcolsep{5pt}
    \begin{tabular}{l|rrrr}
    \toprule
     & \multicolumn{1}{c}{JDDC} & \multicolumn{1}{c}{SGD} & \multicolumn{1}{c}{ReDial} & \multicolumn{1}{c}{CCPE} \\ \midrule
    HiGRU \citep{10.1145/3404835.3463241} & 17.1 & 8.6 & 8.3 & 27.4 \\ 
    BERT \citep{10.1145/3404835.3463241} & 18.5 & 4.8 & 12.5 & 24.5 \\
    SatAct \citep{10.1145/3477495.3531814} & - & 71.3 & - & 16.5 \\
    SatActUtt \citep{10.1145/3477495.3531814} & - & \textbf{84.7} & - & 73.4 \\
    ContextBERT-ERToD (0-shot) & \textbf{50.8} & 78.8 & \textbf{78.1} & \textbf{77.6} \\
    \bottomrule
    \end{tabular} 
    \caption{\label{tab:0shot-uss} Binary F1 scores on different USS subsets. Best scores are marked in \textbf{bold}.}
\vspace{-3mm}
\end{table}

Following existing work, we first report binary F1 for direct comparison. In Table \ref{tab:0shot-uss}, ContextBERT-ERToD performs comparably with SatActUtt and significantly outperforms other models. This shows that our ERToD framework in combination with the ERC model generalises well to user satisfaction prediction. 

\section{Conclusion}
In this work, we propose ERToD, a framework to address three critical steps in learning and effectively adapt chit-chat ERC models to recognise emotions in ToDs. We propose two strategies of DA for different emotions to improve ERC performance in ToDs on rare emotions. We further leverage dialogue state and sentiment-aware embeddings for a richer feature representation. In addition, we apply MTL and devise a novel loss function, \textit{EmoDistLoss}, which take the similarities between emotions into account. Our framework significantly improves existing chit-chat ERC models' performance in recognising user emotions in ToDs. By further applying our best resulting model to perform the task of user satisfaction prediction, we show that our method generalises well on other similar valence-related classification tasks in ToDs.

As more sophisticated and powerful dialogue systems such as ChatGPT arise, there is an urge to recognise, understand and handle the emotion of the user, especially in the age where online abuse is omnipresent. The long-term aim of this work is to obtain valuable insight for downstream ToD modelling tasks. This allows further investigation of emotion regulation strategies on the system side to improve task performance and user satisfaction, and to prevent undesirable user behaviours. 

\section{Acknowledgements}
S. Feng, N. Lubis, M. Heck, and C. van Niekerk are supported by funding provided by the Alexander von Humboldt Foundation in the framework of the Sofja Kovalevskaja Award endowed by the Federal Ministry of Education and Research, while C. Geishauser, H-C. Lin, B. Ruppik, and R. Vukovic are supported by funds from the European Research Council (ERC) provided under the Horizon 2020 research and innovation programme (Grant agreement No. STG2018804636). Computing resources were provided by Google Cloud.

\bibliography{anthology,custom}
\bibliographystyle{acl_natbib}

\appendix

\newpage
\onecolumn
\setcounter{table}{0}
\counterwithin{figure}{section}
\renewcommand{\thetable}{\Alph{section}\arabic{table}}

\section{Emotion Definitions in EmoWOZ}
\label{sec:emotion-definition}
\begin{table}[H]
\centering
\scriptsize
\setlength\tabcolsep{2pt}
\renewcommand{\arraystretch}{1.1}
\begin{tabular}{ccclll}
\toprule[1pt]
\textbf{Elicitor} & \textbf{Valence} & \textbf{Conduct} & \multicolumn{1}{c}{\textbf{OCC Emotion Tokens}} & \multicolumn{1}{c}{\textbf{EmoWOZ Emotion}} & \multicolumn{1}{c}{\textbf{Implication of User}} \\ \hline
 &  & Polite &  & {\color[HTML]{000000} \textcolor{blue}{Satisfied}, liking, appreciative} & Satisfied with the operator because the goal is fulfilled. \\ \cline{3-3} \cline{5-6} 
 & \multirow{-2}{*}{Positive} & Impolite & \multirow{-2}{*}{Admiration, gratitude, love} & \multicolumn{2}{l}{{\color[HTML]{9B9B9B} Not applicable to EmoWOZ}} \\ \cline{2-6} 
 &  & Polite &  & \textcolor{blue}{Dissatisfied}, disliking & Dissatisfied with the operator's suggestion or mistake. \\ \cline{3-3} \cline{5-6} 
\multirow{-4}{*}{Operator} & \multirow{-2}{*}{Negative} & Impolite & \multirow{-2}{*}{Reproach, anger, hate} & \textcolor{blue}{Abusive} & Insulting the operator when the goal is not fulfilled. \\ \hline
 &  & Polite &  & \multicolumn{2}{l}{{\color[HTML]{9B9B9B} }} \\ \cline{3-3}
 & \multirow{-2}{*}{Positive} & Impolite & \multirow{-2}{*}{Pride, gratification} & \multicolumn{2}{l}{\multirow{-2}{*}{{\color[HTML]{9B9B9B} Not applicable to EmoWOZ}}} \\ \cline{2-6} 
 &  & Polite &  & \textcolor{blue}{Apologetic} & Apologising for causing confusion to the operator. \\ \cline{3-3} \cline{5-6} 
\multirow{-4}{*}{User} & \multirow{-2}{*}{Negative} & Impolite & \multirow{-2}{*}{Shame, remorse, hate} & {\color[HTML]{9B9B9B} Not modelled in EmoWOZ} & {\color[HTML]{9B9B9B} Insulting the operator for no reason.} \\ \hline
 &  & Polite &  & \textcolor{blue}{Excited}, happy, anticipating & Looking forward to a good event (e.g. birthday party). \\ \cline{3-3} \cline{5-6} 
 & \multirow{-2}{*}{Positive} & Impolite & \multirow{-2}{*}{\begin{tabular}[c]{@{}l@{}}Happy-for, gloating, love,\\ satisfaction, relief, joy\end{tabular}} & \multicolumn{2}{l}{{\color[HTML]{9B9B9B} Not applicable to EmoWOZ}} \\ \cline{2-6} 
 &  & Polite &  & \textcolor{blue}{Fearful}, sad, disappointed & Encountered a bad event (e.g. robbery and option not available). \\ \cline{3-3} \cline{5-6} 
\multirow{-4}{*}{\begin{tabular}[c]{@{}c@{}}Events,\\ facts\end{tabular}} & \multirow{-2}{*}{Negative} & Impolite & \multirow{-2}{*}{\begin{tabular}[c]{@{}l@{}}Distress, resentment, hate, fears-\\ confirmed, pity, disappointment\end{tabular}} & \multicolumn{2}{l}{\color[HTML]{9B9B9B} Not applicable to EmoWOZ} \\ \hline
 &  & Polite &  & \textcolor{blue}{Neutral} & Describing situations and needs. \\ \cline{3-3} \cline{5-6} 
\multirow{-2}{*}{-} & \multirow{-2}{*}{Neutral} & Impolite & \multirow{-2}{*}{-} & {\color[HTML]{9B9B9B} Not modelled in EmoWOZ} & {\color[HTML]{9B9B9B} No emotion but rude (e.g. using imperative sentences).} \\
\bottomrule[1pt]
\end{tabular}
\caption{EmoWOZ labels and similar emotions tokens from the OCC emotion model. For simplicity, emotion words in blue are used to represent each emotion category.}
\label{tab:our-emotions}

\end{table}

\section{Examples of Augmented Samples}
\label{sec:da-examples}
\subsection{Augmentation with Automatic Label}

\begin{figure}[H]
\centering
\includegraphics[width=\textwidth]{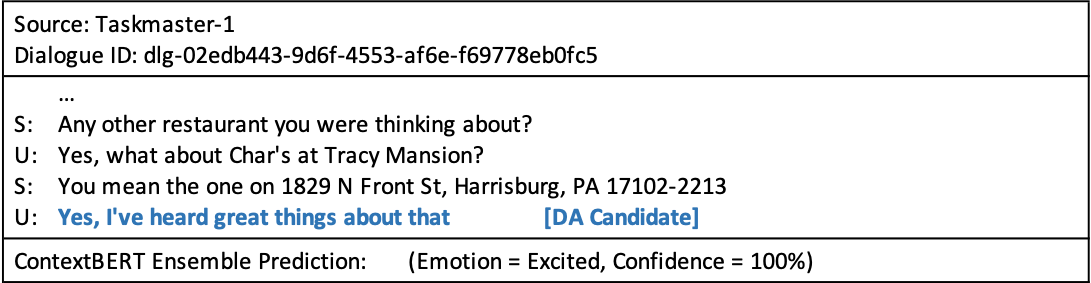}
\caption{DA sample for emotion \textit{Excited}.\label{fig:da-excited}}
\end{figure}

\begin{figure}[H]
\centering
\includegraphics[width=\textwidth]{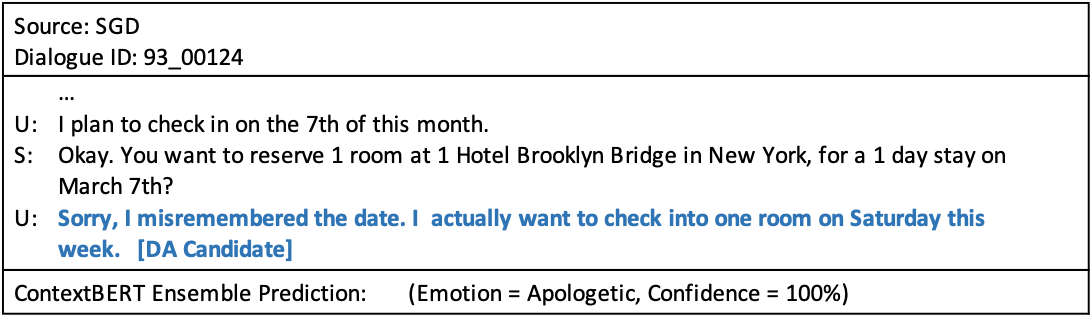}
\caption{DA sample for emotion \textit{Apologetic}.\label{fig:da-apologetic}}
\end{figure}

\begin{figure}[H]
\centering
\includegraphics[width=\textwidth]{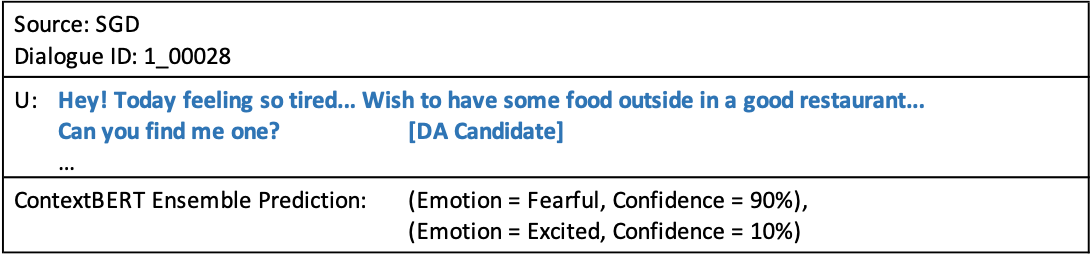}
\caption{DA sample for emotion \textit{Fearful}. Please note that although this class is called ``fearful'' for simplicity, user's negative emotion due to any undesirable events that is out of the control of the operator also belongs to this category in EmoWOZ according to Table \ref{tab:our-emotions}. \label{fig:da-fearful}}
\end{figure}

\subsection{Augmentation with Existing Dataset and Utterance Replacement}
\begin{figure}[H]
\centering
\includegraphics[width=\textwidth]{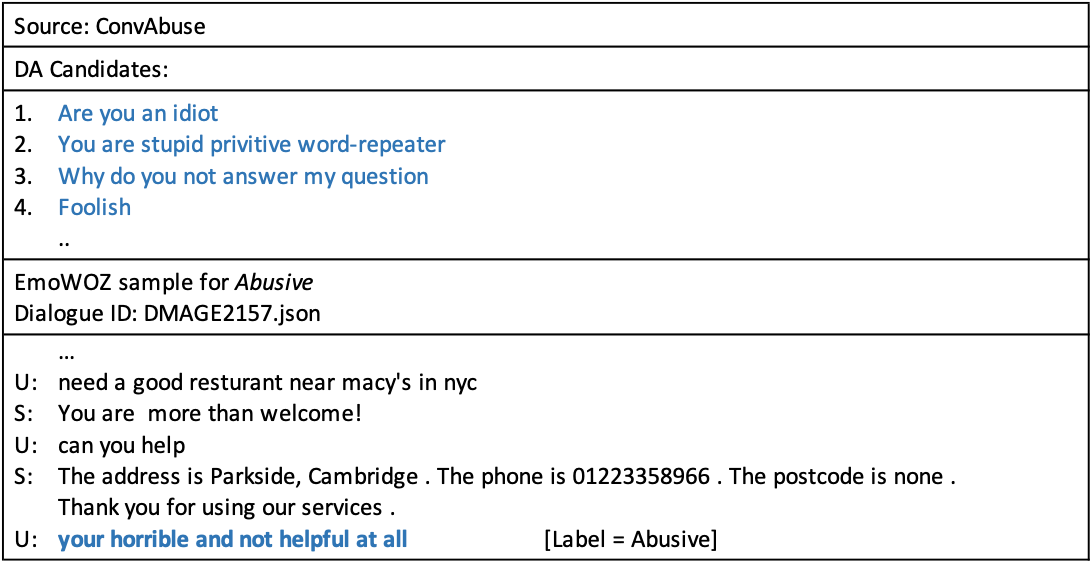}
\caption{DA sample for emotion \textit{Abusive}. Candidate DA samples from ConvAbuse can be used to replace the abusive user utterance in EmoWOZ without causing any context inconsistency.\label{fig:da-abusive}}
\end{figure}

\section{Emotional Aspect Distance Definition}
\label{sec:emotion-distance-definition}
\begin{figure}[H]
\centering
\includegraphics[width=\columnwidth]{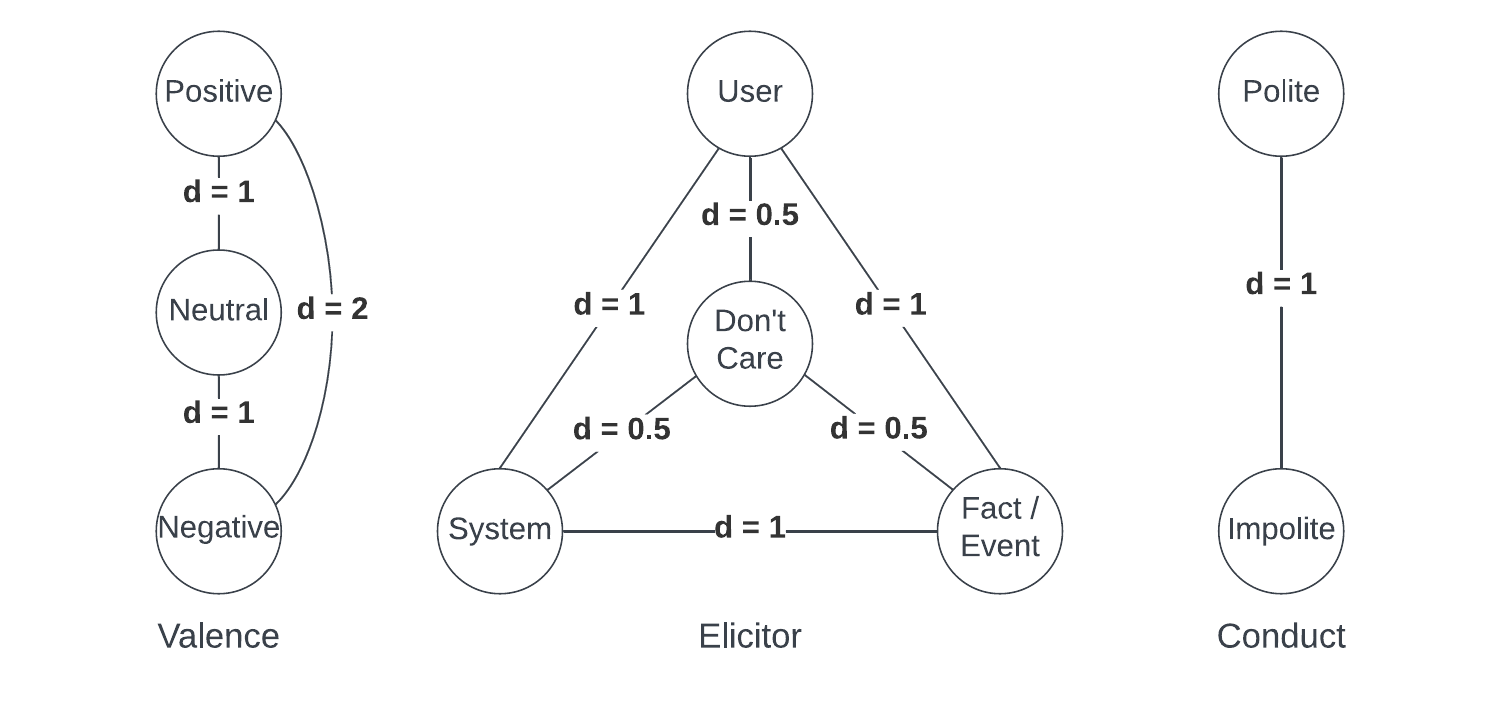}
\caption{Distance definition for Equation \ref{eqn:dist-definition}.\label{fig:distance}}
\end{figure}

\section{Examples of Model Recognitions}
\label{sec:model-predictions}
\begin{figure}[H]
\centering
\includegraphics[width=\textwidth]{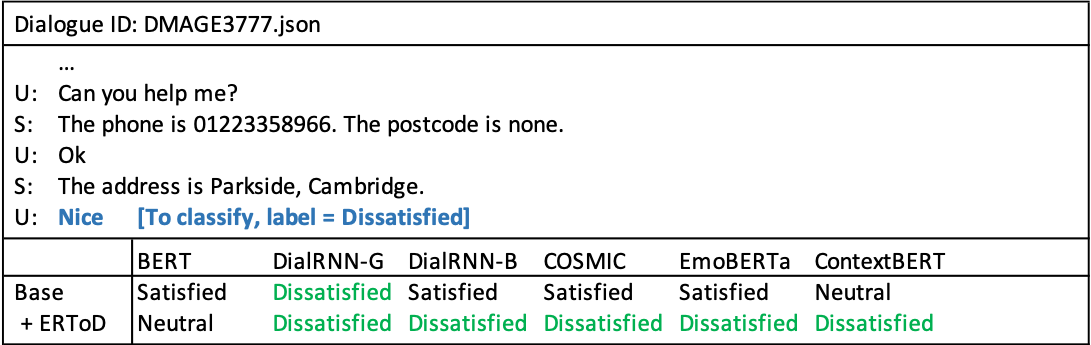}
\caption{Model Recognitions on dialogue DMAGE3777 in EmoWOZ.\label{fig:dmage-3777}}
\end{figure}

\begin{figure}[H]
\centering
\includegraphics[width=\textwidth]{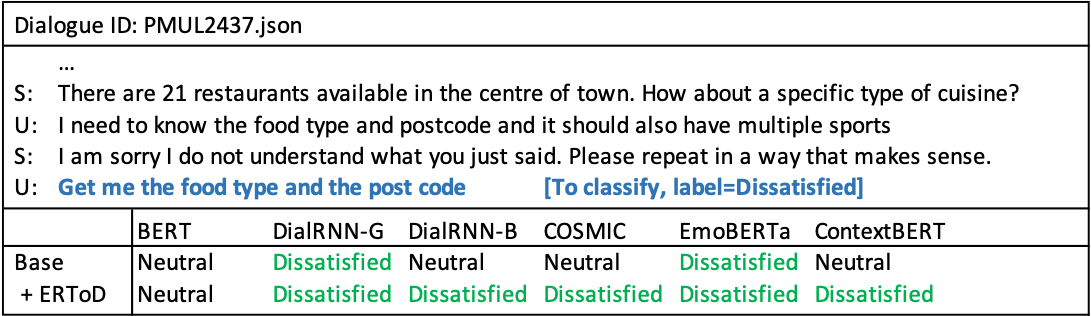}
\caption{Model Recognitions on dialogue PMUL2437 in EmoWOZ\label{fig:pmul-2437}}
\end{figure}

\section{Detailed ERC Performance on Each Emotion}
\label{sec:f1-all-emotions}

\begin{table}[H]
\centering
\setlength\tabcolsep{5.4pt}
\begin{tabular}{l|r|r|r|r|r|r|r}
\toprule[1pt]
\multicolumn{1}{c|}{Model} & \multicolumn{1}{l|}{Neutral} & \multicolumn{1}{l|}{Satisfied} & \multicolumn{1}{l|}{Dissatisfied} & \multicolumn{1}{l|}{Excited} & \multicolumn{1}{l|}{Apologetic} & \multicolumn{1}{l|}{Fearful} & \multicolumn{1}{l}{Abusive}  \\ \midrule
BERT & 89.8 & 88.8 & 35.1 & 42.9 & 70.4 & 36.2 & 27.5  \\
DialogueRNN+GloVe & 83.5 & 86.4 & 51.4 & 32.7 & 57.7 & 12.7 & 0.0  \\
DialogueRNN+BERT & 86.9 & 87.6 & 47.5 & 39.4 & \textbf{71.5} & 41.3 & 25.6  \\
COSMIC & 89.8 & 88.4 & 50.7 & 44.4 & 70.9 & \textbf{52.0} & 31.6  \\
EmoBERTa & \textbf{94.0} & \textbf{90.3} & \textbf{71.0} & 44.9 & 70.6 & 31.3 & 39.3 \\ 
ContextBERT & 93.5 & 89.1 & 69.7 & \textbf{45.6} & 69.6 & 33.3 & \textbf{47.0} \\
\bottomrule
\end{tabular}
\caption{F1 scores of selected chit-chat ERC models BEFORE incorporating ERToD framework. The best score for each emotion is marked in \textbf{bold}.
\label{tab:f1-emotion-base-models}}
\end{table}

\begin{table}[H]
\centering
\setlength\tabcolsep{2.9pt}
\footnotesize
\begin{tabular}{l|rr|rr|rr|rr|rr|rr|rr|rr|rr}
\toprule
 & \multicolumn{2}{c|}{Neu.} & \multicolumn{2}{c|}{Sat.} & \multicolumn{2}{c|}{Dis.} & \multicolumn{2}{c|}{Exc.} & \multicolumn{2}{c|}{Apo.} & \multicolumn{2}{c|}{Fea.} & \multicolumn{2}{c|}{Abu.} & \multicolumn{2}{c|}{M-Avg} & \multicolumn{2}{c}{W-Avg} \\
 & \multicolumn{1}{c}{P} & \multicolumn{1}{c|}{R} & \multicolumn{1}{c}{P} & \multicolumn{1}{c|}{R} & \multicolumn{1}{c}{P} & \multicolumn{1}{c|}{R} & \multicolumn{1}{c}{P} & \multicolumn{1}{c|}{R} & \multicolumn{1}{c}{P} & \multicolumn{1}{c|}{R} & \multicolumn{1}{c}{P} & \multicolumn{1}{c|}{R} & \multicolumn{1}{c}{P} & \multicolumn{1}{c|}{R} & \multicolumn{1}{c}{P} & \multicolumn{1}{c|}{R} & \multicolumn{1}{c}{P} & \multicolumn{1}{c}{R} \\ \midrule
BERT & 90.3 & 89.3 & 88.4 & 89.2 & 38.9 & 38.6 & \textbf{47.7} & 39.1 & 69.7 & 71.5 & 47.7 & 30.0 & 42.1 & 22.4 & 55.7 & 48.5 & 74.5 & 74.5 \\
DialRNN-GloVe & \textbf{97.6} & 73.0 & 78.5 & \textbf{95.9} & 36.5 & \textbf{87.6} & 22.2 & \textbf{65.7} & 44.7 & \textbf{82.5} & 11.2 & 18.9 & 0 & 0 & 32.2 & \textbf{58.4} & 65.0 & \textbf{91.4} \\
DialRNN-BERT & 94.0 & 80.7 & 84.7 & 90.7 & 34.8 & 75.3 & 36.5 & 42.9 & 68.3 & 75.0 & 46.7 & 37.5 & 28.6 & 23.5 & 49.9 & 57.5 & 70.4 & 84.2 \\
COSMIC & 93.1 & 86.8 & 86.2 & 90.7 & 42.3 & 64.4 & 43.7 & 45.3 & \textbf{71.9} & 70.1 & \textbf{65.0} & \textbf{43.3} & \textbf{77.3} & 20.0 & \textbf{64.4} & 55.6 & 74.0 & 81.7 \\
EmoBERTa & 94.2 & \textbf{94.0} & \textbf{88.7} & 92.2 & \textbf{74.6} & 69.5 & 45.6 & 42.6 & 73.0 & 70.3 & 37.9 & 27.2 & 54.0 & 24.7 & 62.3 & 54.4 & \textbf{82.9} & 83.8 \\
ContextBERT & 93.4 & 93.7 & 88.5 & 89.8 & 72.6 & 67.2 & 46.4 & 45.4 & 68.3 & 71.6 & 37.9 & 30.0 & 64.5 & \textbf{37.6} & 63.0 & 57.0 & 82.3 & 81.8 \\
\bottomrule
\end{tabular}
\caption{\textbf{P}recision and \textbf{R}ecall scores of selected chit-chat ERC models BEFORE incorporating ERToD framework. We report scores of each emotion: \textbf{Neu}tral, \textbf{Sat}isfied, \textbf{Dis}satisfied, \textbf{Exc}ited, \textbf{Apo}logetic, \textbf{Fea}rful, \textbf{Abu}sive, as well as \textbf{M}acro- and \textbf{W}eighted \textbf{Av}era\textbf{g}ed scores. The best score for each emotion is marked in \textbf{bold}. Neutral is excluded when calculating the averaged scores. For better presentation, DialogueRNN is shortened to DialRNN.
\label{tab:precision-recall-all-before}}
\end{table}

\begin{table}[H]
\centering
\setlength\tabcolsep{5.4pt}
\begin{tabular}{l|r|r|r|r|r|r|r}
\toprule[1pt]
\multicolumn{1}{c|}{Model} & \multicolumn{1}{l|}{Neutral} & \multicolumn{1}{l|}{Satisfied} & \multicolumn{1}{l|}{Dissatisfied} & \multicolumn{1}{l|}{Excited} & \multicolumn{1}{l|}{Apologetic} & \multicolumn{1}{l|}{Fearful} & \multicolumn{1}{l}{Abusive}  \\ \midrule
BERT & 92.4 & 90.4 & 43.7 & \textbf{49.7} & 75.4 & 39.5 & \textbf{69.7} \\ 
DialogueRNN+GloVe & 92.6 & 90.1 & 51.4 & 43.9 & \textbf{77.6} & 42.4 & 33.8 \\
DialogueRNN+BERT & 92.6 & 90.1 & 51.4 & 43.9 & \textbf{77.6} & 42.4 & 33.8 \\
COSMIC & 91.1 & 89.5 & 58.1 & 45.6 & 73.3 & 36.3 & 41.6 \\
EmoBERTa & \textbf{94.0} & \textbf{90.5} & \textbf{72.3} & 47.9 & 71.9 & \textbf{43.4} & \textbf{69.7} \\ 
ContextBERT & \textbf{94.0} & \textbf{90.5} & \textbf{72.3} & 47.9 & 71.9 & \textbf{43.4} & \textbf{69.7} \\
\bottomrule
\end{tabular}
\caption{F1 scores of selected chit-chat ERC models AFTER incorporating ERToD framework. The best score for each emotion is marked in \textbf{bold}.
\label{tab:f1-emotion-ertod}}
\end{table}

\begin{table}[H]
\centering
\setlength\tabcolsep{2.9pt}
\footnotesize
\begin{tabular}{l|rr|rr|rr|rr|rr|rr|rr|rr|rr}
\toprule
 & \multicolumn{2}{c|}{Neu.} & \multicolumn{2}{c|}{Sat.} & \multicolumn{2}{c|}{Dis.} & \multicolumn{2}{c|}{Exc.} & \multicolumn{2}{c|}{Apo.} & \multicolumn{2}{c|}{Fea.} & \multicolumn{2}{c|}{Abu.} & \multicolumn{2}{c|}{M-Avg} & \multicolumn{2}{c}{W-Avg} \\
 & \multicolumn{1}{c}{P} & \multicolumn{1}{c|}{R} & \multicolumn{1}{c}{P} & \multicolumn{1}{c|}{R} & \multicolumn{1}{c}{P} & \multicolumn{1}{c|}{R} & \multicolumn{1}{c}{P} & \multicolumn{1}{c|}{R} & \multicolumn{1}{c}{P} & \multicolumn{1}{c|}{R} & \multicolumn{1}{c}{P} & \multicolumn{1}{c|}{R} & \multicolumn{1}{c}{P} & \multicolumn{1}{c|}{R} & \multicolumn{1}{c}{P} & \multicolumn{1}{c|}{R} & \multicolumn{1}{c}{P} & \multicolumn{1}{c}{R} \\ \midrule
BERT & 91.0 & 93.8 & 88.9 & 92.0 & 57.5 & 35.5 & \textbf{51.2} & 48.9 & \textbf{81.6} & 70.3 & 48.1 & 33.9 & \textbf{74.8} & 65.9 & 67.0 & 57.7 & 79.8 & 76.3 \\
DialRNN-GloVe & 91.3 & \textbf{94.0} & \textbf{89.7} & 90.5 & 60.9 & 41.5 & 44.4 & 45.6 & 76.5 & 77.3 & 42.6 & \textbf{38.3} & 54.3 & 30.0 & 61.4 & 53.9 & 80.6 & 76.5 \\
DialRNN-BERT & 91.3 & \textbf{94.0} & \textbf{89.7} & 90.5 & 60.9 & 41.5 & 44.4 & 45.6 & 76.5 & 77.3 & 42.6 & \textbf{38.3} & 54.3 & 30.0 & 61.4 & 53.9 & 80.6 & 76.5 \\
COSMIC & \textbf{94.4} & 88.3 & 86.9 & 92.3 & 51.6 & \textbf{68.9} & 38.7 & \textbf{57.4} & 68.2 & \textbf{79.3} & 36.2 & \textbf{38.3} & 44.7 & 38.8 & 54.4 & 62.5 & 75.9 & \textbf{84.6} \\
EmoBERTa & 94.3 & 93.9 & 88.9 & \textbf{92.4} & \textbf{75.6} & 68.0 & 45.7 & 50.7 & 70.8 & 74.4 & \textbf{54.6} & 35.6 & 72.4 & \textbf{68.2} & \textbf{68.0} & \textbf{64.9} & \textbf{83.5} & 84.3 \\
ContextBERT & 94.3 & 93.9 & 88.9 & \textbf{92.4} & \textbf{75.6} & 68.0 & 45.7 & 50.7 & 70.8 & 74.4 & \textbf{54.6} & 35.6 & 72.4 & \textbf{68.2} & \textbf{68.0} & \textbf{64.9} & \textbf{83.5} & 84.3 \\
\bottomrule
\end{tabular}
\caption{\textbf{P}recision and \textbf{R}ecall scores of selected chit-chat ERC models AFTER incorporating ERToD framework. We report scores of each emotion: \textbf{Neu}tral, \textbf{Sat}isfied, \textbf{Dis}satisfied, \textbf{Exc}ited, \textbf{Apo}logetic, \textbf{Fea}rful, \textbf{Abu}sive, as well as \textbf{M}acro- and \textbf{W}eighted \textbf{Av}era\textbf{g}ed scores. The best score for each emotion is marked in \textbf{bold}. Neutral is excluded when calculating the averaged scores. For better presentation, DialogueRNN is shortened to DialRNN.
\label{tab:precision-recall-all-after}}
\end{table}

\begin{table}[H]
\centering
\setlength\tabcolsep{5.4pt}
\begin{tabular}{l|r|r|r|r|r|r|r}
\toprule[1pt]
\multicolumn{1}{c|}{Model} & \multicolumn{1}{l|}{Neutral} & \multicolumn{1}{l|}{Satisfied} & \multicolumn{1}{l|}{Dissatisfied} & \multicolumn{1}{l|}{Excited} & \multicolumn{1}{l|}{Apologetic} & \multicolumn{1}{l|}{Fearful} & \multicolumn{1}{l}{Abusive}  \\ \midrule
BERT & +2.6 & +1.6 & +8.6 & +6.8 & +5.0 & +3.3 & +42.2  \\
DialogueRNN+GloVe & +9.1 & +3.7 & +0.0 & +11.2 & +19.9 & +29.7 & +33.8 \\
DialogueRNN+BERT & +5.7 & +2.5 & +3.9 & +4.5 & +6.1 & +1.1 & +8.2 \\
COSMIC & +1.3 & +1.1 & +7.4 & +1.2 & +2.4 & \textbf{-15.7} & +10.0 \\
EmoBERTa & 0.0 & +0.2 & +1.3 & +3.0 & +1.3 & +12.1 & +30.4 \\ 
ContextBERT & +0.5 & +1.4 & +2.6 & +2.3 & +2.3 & +10.1 & +22.7 \\
\bottomrule
\end{tabular}
\caption{Change of F1 scores of selected chit-chat ERC models after incorporating ERToD framework. The only degradation in performance is marked in \textbf{bold}.
\label{tab:f1-emotion-difference}}
\end{table}

In terms of F1 scores, ERToD results in improvement in all emotions except for \textit{fearful} in COSMIC (Table \ref{tab:f1-emotion-difference}). We further investigate this exception. 
While most of fearful utterances are located at the beginning of the dialogue in the training and development set in EmoWOZ, the position of such utterances are more evenly distributed in the test set as well as the augmented samples. 
Upon toggling the development set and the test set for evaluation, we observe that the F1 of fearful by COSMIC drops significantly (52.0\% $\rightarrow$ 28.8\%) while that of COSMIC-ERToD remains roughly unchanged (35.5\% $\rightarrow$ 37.6\%). The trend in all other results remains unchanged.

The drastically different performance of COSMIC on the development and the test set suggests that COSMIC develops a positional bias from the training set of EmoWOZ. At the same time, COSMIC-ERToD performs similarly on both non-training sets, likely relying more on textual and task information. The limited performance of COSMIC-ERToD is likely due to the extra false-positives at the later stage of dialogues.

\begin{table}[H]
\centering
\setlength\tabcolsep{1pt}
\footnotesize
\begin{tabular}{l|rr|rr|rr|rr|rr|rr|rr|rr|rr}
\toprule
 & \multicolumn{2}{c|}{Neu.} & \multicolumn{2}{c|}{Sat.} & \multicolumn{2}{c|}{Dis.} & \multicolumn{2}{c|}{Exc.} & \multicolumn{2}{c|}{Apo.} & \multicolumn{2}{c|}{Fea.} & \multicolumn{2}{c|}{Abu.} & \multicolumn{2}{c|}{M-Avg} & \multicolumn{2}{c}{W-Avg} \\
 & \multicolumn{1}{c}{P} & \multicolumn{1}{c|}{R} & \multicolumn{1}{c}{P} & \multicolumn{1}{c|}{R} & \multicolumn{1}{c}{P} & \multicolumn{1}{c|}{R} & \multicolumn{1}{c}{P} & \multicolumn{1}{c|}{R} & \multicolumn{1}{c}{P} & \multicolumn{1}{c|}{R} & \multicolumn{1}{c}{P} & \multicolumn{1}{c|}{R} & \multicolumn{1}{c}{P} & \multicolumn{1}{c|}{R} & \multicolumn{1}{c}{P} & \multicolumn{1}{c|}{R} & \multicolumn{1}{c}{P} & \multicolumn{1}{c}{R} \\ \midrule
BERT & +0.7 & +4.5 & +0.5 & +2.8 & +18.6 & -3.1 & +3.5 & +9.8 & +11.9 & -1.2 & +0.4 & +3.9 & +32.7 & +43.5 & +11.3 & +9.2 & +5.3 & +1.8 \\
DialRNN-GloVe & -6.3 & +21.0 & +11.2 & -5.4 & +24.4 & -46.1 & +22.2 & -20.1 & +31.8 & -5.2 & +31.4 & +19.4 & +54.3 & +30.0 & +29.2 & -4.5 & +15.6 & -14.9 \\
DialRNN-BERT & -2.7 & +13.3 & +5.0 & -0.2 & +26.1 & -33.8 & +7.9 & +2.7 & +8.2 & +2.3 & -4.1 & +0.8 & +25.7 & +6.5 & +11.5 & -3.6 & +10.2 & -7.7 \\
COSMIC & +1.3 & +1.5 & +0.7 & +1.6 & +9.3 & +4.5 & -5.0 & +12.1 & -3.7 & +9.2 & -28.8 & -5.0 & -32.6 & +18.8 & -10.0 & +6.9 & +1.9 & +2.9 \\
EmoBERTa & +0.1 & -0.1 & +0.2 & +0.2 & +1.0 & -1.5 & +0.1 & +8.1 & -2.2 & +4.1 & +16.7 & +8.4 & +18.4 & +43.5 & +5.7 & +10.5 & +0.6 & +0.5 \\
ContextBERT & +0.9 & +0.2 & +0.4 & +2.6 & +3.0 & +0.8 & -0.7 & +5.3 & +2.5 & +2.8 & +16.7 & +5.6 & +7.9 & +30.6 & +5.0 & +7.9 & +1.2 & +2.5 \\
\bottomrule
\end{tabular}
\caption{The difference in \textbf{P}recision and \textbf{R}ecall scores of selected chit-chat ERC models before and after incorporating ERToD framework. We report scores of each emotion: \textbf{Neu}tral, \textbf{Sat}isfied, \textbf{Dis}satisfied, \textbf{Exc}ited, \textbf{Apo}logetic, \textbf{Fea}rful, \textbf{Abu}sive, as well as \textbf{M}acro- and \textbf{W}eighted \textbf{Av}era\textbf{g}ed scores. The best score for each emotion is marked in \textbf{bold}. Neutral is excluded when calculating the averaged scores. For better presentation, DialogueRNN is shortened to DialRNN.
\label{tab:precision-recall-all-diff}}
\end{table}

\section{Averaged Scores for the Ablation Study}
\label{sec:f1-ablation-average}

\begin{table}[H]
\centering
\begin{tabular}{c|l|r|r}
\toprule[1pt]
 & \multicolumn{1}{c|}{Model} & Macro Avg & Weighted Avg \\ \midrule
\multicolumn{1}{l|}{\multirow{6}{*}{\rotatebox[origin=c]{90}{F1 Score ($\uparrow$)}}} & ContextBERT & 59.1 & 81.9 \\
\multicolumn{1}{l|}{} &  \ + DA & $\dagger$64.1 & $\dagger$83.4 \\
\multicolumn{1}{l|}{} &  \ \ + DS & $\dagger$64.1 & $\dagger$83.5 \\
\multicolumn{1}{l|}{} &  \ \ \ + SentiX & $\dagger$64.8 & $\dagger$83.7 \\
\multicolumn{1}{l|}{} &  \ \ \ \ + MTL & $\dagger$65.3 & $\dagger$83.7 \\
\multicolumn{1}{l|}{} & \ + ERToD & $\dagger$\textbf{65.7} & $\dagger$\textbf{83.9} \\ \midrule
\multicolumn{1}{l|}{\multirow{6}{*}{\rotatebox[origin=c]{90}{AED Score ($\downarrow$)}}} & ContextBERT & 0.387 & 0.168 \\ 
\multicolumn{1}{l|}{} & \ + DA & $\dagger$0.351 & $\dagger$0.159 \\ 
\multicolumn{1}{l|}{} & \ \ + DS & $\dagger$0.335 & $\dagger$0.151 \\
\multicolumn{1}{l|}{} & \ \ \ + SentiX & $\dagger$0.331 & $\dagger$0.149 \\
\multicolumn{1}{l|}{} & \ \ \ \ + MTL & $\dagger$0.322 & $\dagger$0.147 \\
\multicolumn{1}{l|}{} &  \ + ERToD  & $\dagger$\textbf{0.316} & $\dagger$\textbf{0.145}\\
\bottomrule
\end{tabular}

\caption{Ablation Study of ERToD. $\dagger$ indicates statistically significant difference with $p<0.05$ when comparing with ContextBERT. The best score in each category is in \textbf{bold}. For each of the additional methods: DA = Data Augmentation, DS = Dialogue State Features, SentiX = Sentiment-aware Text Embedding, MTL = Multi-task Learning. Neutral is excluded when calculating the averaged scores.
\label{tab:ablation-average}}
\end{table}

\begin{table}[H]
\centering
\setlength\tabcolsep{2.9pt}
\footnotesize
\begin{tabular}{l|rr|rr|rr|rr|rr|rr|rr|rr|rr}
\toprule
 & \multicolumn{2}{c|}{Neu.} & \multicolumn{2}{c|}{Sat.} & \multicolumn{2}{c|}{Dis.} & \multicolumn{2}{c|}{Exc.} & \multicolumn{2}{c|}{Apo.} & \multicolumn{2}{c|}{Fea.} & \multicolumn{2}{c|}{Abu.} & \multicolumn{2}{c|}{M-Avg} & \multicolumn{2}{c}{W-Avg} \\
 & \multicolumn{1}{c}{P} & \multicolumn{1}{c|}{R} & \multicolumn{1}{c}{P} & \multicolumn{1}{c|}{R} & \multicolumn{1}{c}{P} & \multicolumn{1}{c|}{R} & \multicolumn{1}{c}{P} & \multicolumn{1}{c|}{R} & \multicolumn{1}{c}{P} & \multicolumn{1}{c|}{R} & \multicolumn{1}{c}{P} & \multicolumn{1}{c|}{R} & \multicolumn{1}{c}{P} & \multicolumn{1}{c|}{R} & \multicolumn{1}{c}{P} & \multicolumn{1}{c|}{R} & \multicolumn{1}{c}{P} & \multicolumn{1}{c}{R} \\ \midrule
ContextBERT & 93.4 & 93.9 & 88.5 & \textbf{92.4} & 72.6 & 68.0 & 46.4 & 50.7 & 68.3 & 74.4 & 37.9 & \textbf{35.6} & 64.5 & 68.2 & 63.0 & 64.9 & 82.3 & 84.3 \\
\ + DA & 93.9 & 94.4 & 89.4 & 91.6 & 75.6 & 67.2 & 47.2 & 44.6 & 75.0 & 70.0 & 53.1 & 30.6 & 70.1 & 65.9 & \textbf{68.4} & 61.6 & 84.0 & 83.1 \\
\ \ + DS & 93.8 & \textbf{94.6} & \textbf{90.1} & 90.9 & 74.5 & 68.4 & \textbf{47.9} & 44.6 & 75.8 & 69.0 & 50.7 & 27.8 & 69.9 & 69.4 & 68.1 & 61.7 & \textbf{84.2} & 82.9 \\
\ \ \ + SentiX & 94.1 & 94.3 & 89.5 & 91.7 & 76.0 & 69.1 & 47.5 & 49.3 & \textbf{76.7} & 70.3 & 50.9 & 32.2 & 66.0 & 66.5 & 67.8 & 63.2 & 84.1 & 83.9 \\
\ \ \ \ + MTL & 94.2 & 94.0 & 88.9 & 91.5 & \textbf{76.4} & \textbf{70.6} & 45.7 & \textbf{49.8} & 76.6 & \textbf{71.6} & 51.2 & 35.0 & 67.0 & \textbf{72.4} & 67.6 & \textbf{65.1} & 83.8 & \textbf{84.2} \\
\ + ERToD & \textbf{94.3} & 94.1 & 88.9 & 91.9 & 75.6 & 69.3 & 45.7 & 48.8 & 70.8 & 70.8 & \textbf{54.6} & 34.4 & \textbf{72.4} & 70.0 & 68.0 & 64.2 & 83.5 & 84.1 \\
\bottomrule
\end{tabular}
\caption{Ablation study on \textbf{P}recision and \textbf{R}ecall scores of ERToD. We report scores of each emotion: \textbf{Neu}tral, \textbf{Sat}isfied, \textbf{Dis}satisfied, \textbf{Exc}ited, \textbf{Apo}logetic, \textbf{Fea}rful, \textbf{Abu}sive, as well as \textbf{M}acro- and \textbf{W}eighted \textbf{Av}era\textbf{g}ed scores. The best score for each emotion is marked in \textbf{bold}. For each of the additional methods: DA = Data Augmentation, DS = DialogueState Features, SentiX = Sentiment-aware Text Embedding, MTL = Multi-task Learning.. Neutral is excluded when calculating averaged scores.
\label{tab:precision-recall-all-ablation}}
\end{table}

\end{document}